\documentclass[letterpaper, 10 pt, conference]{ieeeconf}  

\IEEEoverridecommandlockouts                              

\usepackage{graphics} 
\usepackage{mathptmx} 
\usepackage{times} 
\usepackage{amsmath} 
\usepackage{amssymb}  
\usepackage{todonotes}
\usepackage{graphicx}
\usepackage{subfig}
\usepackage{multirow}
\usepackage{bm}

\newcommand {\dt}[1] { \dfrac{d#1}{dt} }

\title{\LARGE \bf
Trajectory Generation for Quadrotor Based Systems \\using Numerical Optimal Control
}

\author{Mathieu Geisert$^{1}$ and Nicolas Mansard$^{1}$
\thanks{$^{1}$ Laboratory  for
Analysis and Architecture of Systems (LAAS), Centre National de la Recherche Scientifique (CNRS), Toulouse 31077, France {\tt\small mgeisert@laas.fr}, {\tt\small nmansard@laas.fr}}%
}

\begin{document}

\maketitle
\thispagestyle{empty}
\pagestyle{empty}

\begin{abstract}
The recent works on quadrotor have focused on more and more challenging tasks with increasingly complex systems. Systems are often augmented with slung loads, inverted pendulums or arms, and accomplish complex tasks such as going through a window, grasping, throwing or catching. Usually, controllers are designed to accomplish a specific task on a specific system using analytic solutions, so each application needs long preparations. On the other hand, the direct multiple shooting approach is able to solve complex problems without any analytic development, by using off-the-shelf optimization solver. In this paper, we show that this approach is able to solve a wide range of problems relevant to quadrotor systems, from on-line trajectory generation for quadrotors, to going through a window for a quadrotor-and-pendulum system, through manipulation tasks for a aerial manipulator.
\end{abstract}

\section{INTRODUCTION}

During transportation missions, loads are often carried thanks to a cable to avoid aerodynamic perturbations. Various controllers have been designed for this under-actuated system. For tasks involving only transportation, a typical approach is to reduce the energy of the load to limit its swing motion. Swing-free motion can be found using dynamic programming approach \cite{C6},\cite{C8} or reinforcement learning \cite{C5},\cite{C9}. In \cite{C10},\cite{C11} the differential flatness of the system is used to build a cascade of controllers capable of controlling the load position and attitude. 
In \cite{C}, dynamic programming is used to calculate aggressive trajectories. Using this technique, authors showed that their method is able to find trajectories to go through a window with a system quadrotor and slung load. However the problem of finding how the quadrotor and the load have to pass through the window is solved before hand and given to the algorithm in the form of way-points. In \cite{C2}, trajectory of quadrotors throwing and catching an inverted pendulum are generated off-line using direct optimal control then executed on the real system using a LQR controllers. 

For tasks such as screwing, assembling or other manipulation tasks, robotic arms can be mounted on the flying vehicle. In \cite{QA-G}, a simple symmetric manipulator is used so quadrotor and manipulator controllers can be built apart. When the manipulator becomes more complex, the quadrotor controller has to take into account movements of the arm \cite{QA-C}, \cite{QA-B}, \cite{QA-E}. Once controllers are designed, the problem shift to the trajectory generation. In \cite{QA-A}, \emph{Inverse Kinematics} is used to determine trajectories of the joints and the quadrotor from the end-effector trajectory. In most recent works, optimal control is used to generate on-line picking trajectories \cite{GaKo2015}. The algorithm used is able to exploit the full system dynamics and generate simultaneous trajectories for the quadrotor and the arm but, because it uses an algorithm similar to \emph{Differential Dynamic Programming} (DDP), it cannot handle obstacle constraints well.

In the expectancy to solve problems involving more and more complex tasks on aerial systems, we believe that analytic solutions are too restraining and too difficult to exhibit on a generic manner. An alternative approach, that has not been extensively explored yet, 
is to rather rely on the field of direct optimal control and numerical optimization, to numerically approximate a valid control sequence, that would then be easily adapted to changes in both the task to achieve and the dynamics of the robot.

The contribution of this paper is to show the interest of one of these numerical approaches to design versatile and efficient behaviors on complex aerial vehicles (aerial pendulum and aerial manipulator). We propose a complete formulation to numerically compute an optimal movement. We then report an extensive benchmark of the capabilities of numerical solvers to discover efficient trajectories around obstacles and control these trajectories despite model uncertainties. We also demonstrate that even if those kinds of algorithms can look computationally heavy, they can be used for on-line trajectory generation (\emph{Model Predictive Control}). 

Next section presents the optimal control problem and more specifically, the direct multiple shooting formulation used to solve the problem. Section \ref{sectionSystem} describes the three robot models used for the benchmarks (\emph{i.e} quadrotor, quadrotor with pendulum and quadrotor with robotic arm). Section \ref{sectionImplementationDetails} presents the initial guess, the model of obstacles and the tools used to generate our results. The last section shows the results obtained in simulation for a wide range of tasks.

\section{OPTIMAL CONTROL}

In this section, we briefly recall the framework of optimal control and define explicitly the formulation of the numerical problem that is used for the quadrotors. Meaning while, we justify why we have selected this particular form among all the possibility offered in the literature.

\subsection{Indirect and Direct Approaches}

In this paper, trajectories of different quadrotor based systems are generated using algorithms of the optimal control framework.
The generic optimal control problem can be expressed as follow: 
\begin{equation*}
\underset{\underline{x}\in \mathcal{X}, \underline{u} \in \mathcal{U}}{\text{minimize}} \int_{0}^{T} L(x(t),u(t))dt + E(x(T))
\end{equation*}
subject to $\qquad \forall t\in[0,T]$
\begin{equation*}
\begin{aligned}
&\dot{x}(t) = f(x(t),u(t)), \\
&h(x(t),u(t)) \geq 0,\quad h_{0}(x(0)) \geq 0, \quad h_{T}(x(T)) \geq 0,\\
&r(x(t),u(t)) = 0,\quad r_{0}(x(0))=0, \quad r_{T}(x(T)) = 0. 
\end{aligned}
\end{equation*}
Where the decision variables are the trajectory in state $\underline{x}:t \in [0,T] \to x(t) \in \mathcal{X}$ and in control $\underline{u}:t \in [0,T] \to u(t) \in \mathcal{U}$ (where $\mathcal{X}$ and $\mathcal{U}$ are the state and control spaces, and the underlined symbol is used to differentiate the trajectory from the time value), $L$ represents the integral (or running) cost, $E$ is the terminal cost, $f$ is the system dynamics and the $h$ and $r$ functions represent arbitrary constraints.

Two very different approaches may be considered to solve this problem, the indirect one and the direct one. The indirect one changes the problem into an integration (ODE or DAE) problem using the Pontryagin's maximum principle or the Hamilton-Jacobi-Bellman equation. The resulting problem is a differential equation which is unfortunately often too complex to be integrated as is. When it is possible, this approach provides a complete (and often comparatively cheap) solution to the problem. However, this type of approach is usually applied on a specific system and/or task so the differential equation can be simplify enough to be integrated.  

The direct approach directly solves a discretized approximation of the nominal problem using numerical optimization techniques. It has several advantages: it works directly on the problem so the problem does not need to be reformulated; is solved by generic solvers; and can often hope to directly adapt a solution to variations of both the system dynamics and the tasks.   
This paper only focuses on the direct optimal control. 

\subsection{Direct Approaches}

The discretization of the nominal problem results in a redundant set of decision variables, $\underline{x}$ and $\underline{u}$, constrained by the system dynamics (redundant in the sense the $\underline{x}$ directly arises from $\underline{u}$).
Three different formulations with different properties are typically considered to handle this redundancy.

Single shooting method integrates the whole trajectory from controls using ODE/DAE solvers and optimizes the cost function over controls. This formulation is interesting  because of its simplicity and, since the resulting optimization problem has a low number of degree of freedom and no additional constraints, it easily deals with systems of large dimension \cite{Tassa}.

In colocation method, cost function is optimized over states and control which are linked via the discretized dynamic equation set as constraint for the optimization problem. Those additional constraints force in practice the solver to solve many inverse dynamics problems, which is costly in general, but allow this method to be initialized with state trajectories, to have fast convergence and to deal with unstable systems well \cite{FastMS}. 

Multiple shooting tries to combine advantages of colocation and single shooting by using the same formalism as colocation but with ODE/DAE solvers to integrate the dynamic equation of the system  along time intervals instead of discretizing it. The variables are then a full control trajectory $\underline{u}$ along with a sparse number of state variables $x_{1},..., x_{J}$ called shooting nodes. These nodes avoid the divergence phenomenon encounter in single shooting, and despite an increase in the number of decision variables, speed up the solution \cite{FastMS}.

In this paper, we consider quadrotor based systems: the resulting problem is a small dimension nonlinear problem. Single shooting would be able to calculate each step very quickly however because of the non-linearity and the instability of the system the iterative numerical algorithm would only be able to progress slowly and risks to diverge. Multiple shooting and colocation would probably give similar quality results but thanks to its design where the problem parametrization and the integration of the dynamic are separated, multiple shooting is more flexible and much faster (wich is essential for \emph{Model Predictive Control}). Additionally, the relative stability of quadrotors (compared to rocket or humanoid robot) allows to have only few shooting nodes although non-linearities impose thin steps to integrate the dynamic equations. 
Therefore, the different tasks presented in this paper are solved using the direct multiple shooting approach. 

\subsection{Direct Multiple Shooting}

In multiple shooting methods, the system trajectory is cut into small time intervals that correspond to shooting intervals. A set of $J$ state variables corresponding to the starting point of each shot is introduced.
\begin{equation*}
\begin{aligned}
&x(t_{j}) = x_{j}, && x_{j}\in \mathcal{X}, && j=0,...,J-1 \\
\end{aligned}
\end{equation*}
Thus, integration of the dynamic $\dot{x}(t) = f(x(t),u(t))$ will give a piecewise continuous function with discontinuities at each shooting node that we denote $x(t;x_{j},u_{j})$ for $t_{j} \leq t < t_{j+1}$. Additional constraints are imposed to force continuity at the shooting nodes. Constraints are similarly discretized over time.
\begin{equation*}
\begin{aligned}
&& h(x(t_{j}),u(t_{j}))& \geq 0,\\
&& r(x(t_{j}),u(t_{j}))& = 0, \quad j=0,...,J-1.
\end{aligned}
\end{equation*}
Therefore, the optimal control problem becomes
\begin{equation*}
\underset{x_{0},...,x_{J-1},u_{0},...,u_{J-1} }{\text{minimize}} \sum_{i=0}^{N-1} l_{j}(x_{j},u) + E(x_{N})
\end{equation*}
\quad subject to
\begin{equation*}
\begin{aligned}
x_{j+1} - x(t_{j+1};x_{j},u(t)) = 0,\qquad j = 0,...,J-1, \label{continuityConstraint}\\
\begin{aligned}
h(x(t_{j}),u(t_{j})) \geq 0,&& h_{0}(x(0)) \geq 0,&& h_{T}(x(T)) \geq 0,\nonumber \\
r(x(t_{j}),u(t_{j})) = 0,&& r_{0}(x(0))=0,&& r_{T}(x(T)) = 0. \nonumber
\end{aligned}
\end{aligned}
\end{equation*}
\quad where
\begin{equation*}
l_{j}(x_{j},u) := \int_{t_{j}}^{t_{j+1}} L(x(t;x_{j},u(t)),u(t))dt
\end{equation*}

An other advantage of multiple shooting with respect to single shooting is more technical to get but can be understood from this final formulation. By explicitly express continuity constraints (\ref{continuityConstraint}) over certain points, multiple shooting allows the numerical algorithm to relaxed those constraints. One direct implication is that at the beginning of the algorithm, those constraints can be fully relaxed hence the initial guess does not need to be continuous. Thus, multiple shooting can easily be initialized with other algorithms which give trajectories in the state space and do not take care of the full dynamic like planning algorithm \cite{MS-Planning}. This formulation has also been found useful to control unstable system or to handle terminal constraints because in those cases it can be difficult to find a valid initial guess in the control space. 

The problem generated using direct multiple shooting method can be condensed \cite{Condensing} then be solved using any \emph{NonLinear Programming} (NLP) solver.

\subsection{Sequential Quadratic Programming}

The problem generated using direct multiple shooting can be condensed \cite{Condensing} then be solved using any nonlinear programming solver. In this section, we present the Sequential Quadratic Programming algorithm which is one of the commonly NLP algorithm used to solve this kind of problem. In this part, we consider the following standard-form NLP problem
\begin{equation*}
\underset{x}{\text{min}}\: f(x)\quad \text{subject to} \quad h(x) \geq 0, \quad r(x)=0
\end{equation*}
From this problem, we can introduce the \emph{Lagrangian function}
\begin{equation*}
 \mathcal{L}(x,\lambda,\mu) = f(x) - \lambda^{T}h(x) - \mu^{T}r(x)
 \end{equation*} 
where $\lambda$ and $\mu$ are \emph{Lagrangian multipliers}. The necessary conditions for $x^{\star}$ to be a local optimum of the NLP are that there exist mulpliers $\lambda^{\star}$ and $\mu^{\star}$, such that
\begin{equation*}
\begin{aligned}
&& \nabla \mathcal{L}(x^{\star},\lambda^{\star},\mu^{\star}) &= 0, \\
&& h(x^{\star}) &= 0, \\
&& r(x^{\star}) &\geq 0, \quad \mu^{\star} \geq 0, \quad r(x^{\star})^{T}\mu^{\star}=0.
\end{aligned}
\end{equation*}
At each step of the SQP algorithm, the objective function $f$ is approximate by its local quadratic approximation and constraints by their local affine approximations
\begin{equation*}
\begin{aligned}
f(x) &\approx f(x_{k}) + \nabla f(x_{k})^{T}\Delta x + \frac{1}{2}\Delta x^{T}H_{k}\Delta x \\
h(x) &\approx h(x_{k}) + \nabla h(x_{k})^{T}\Delta x, \\
r(x) &\approx r(x_{k}) + \nabla r(x_{k})^{T}\Delta x. \\
\end{aligned}
\end{equation*}
where $H_{k}$ is a Hessian approximation of $\mathcal{L}(x_{k},\lambda_{k},\mu_{k})$, and $\nabla f(x_{k})$, $\nabla h(x_{k})$ and $\nabla r(x_{k})$ are Jacobians. Thus, at each step of the SQP algorithm, we get the following QP subproblem
\begin{equation*}
\begin{aligned}
&\underset{\Delta x \in \{S,U\}}{\text{minimize}} & \nabla f(x_{k})^{T}\Delta x + \frac{1}{2} \Delta x^{T} H_{k} \Delta x, \\
& \text{subject to} \\
&&h(x_{k}) + \nabla h(x_{k})^{T}\Delta x \geq 0,\\
&&r(x_{k}) + \nabla r(x_{k})^{T}\Delta x= 0. 
\end{aligned}
\end{equation*}
Witch is solved using active-sets methods.
So, starting from an initial guess $(x_{0}, \lambda_{0}, \mu_{0})$, SQP solver decomposed the NLP into a QP subproblem and iterate 
\begin{equation*}
\begin{aligned}
&x_{k+1} =&& x_{k} + \alpha \Delta x^{QP}, \\
& \lambda_{k+1} =&& \lambda_{k} + \alpha \Delta\lambda^{QP}, \\
& \mu_{k+1} =&& \mu_{k} + \alpha \Delta\mu^{QP}. \\ 
\end{aligned}
\end{equation*}
where $\alpha$ can be determined using linesearch methods \cite{NumOpt}.

Constraint linearization can give unsolvable QP subproblems, so in those cases constraints have to be relaxed. One technique is to use $\ell_{1}$ relaxation where the quadratic problem is transformed to
\begin{align}
\underset{\Delta x \in \{S,U\}}{\text{minimize}} &\quad \nabla f(x_{k})^{T}\Delta x + \frac{1}{2} \Delta x^{T} H_{k} \Delta x + \nu u + \nu (v+w), \nonumber \\
 \text{subject to} & \nonumber \\
&h(x_{k}) + \nabla h(x_{k})^{T}\Delta x \geq -u, \label{relaxation}\\
&r(x_{k}) + \nabla r(x_{k})^{T}\Delta x= v-w. 
\end{align}
where $u$, $v$ and $w$ are slack variables and $\nu$ is a penalty parameter. 

SQP solver differ in the way the Hessian is approximated, the linesearch is done, the QP subproblems are solved or the constraints are relaxed. SQP has been shown a powerful tool and because of its superlinear convergence rate and its ability to deal with nonlinear constraints well, it is currently considered as one of the most powerful algorithm to solve large-scale NLP. 

\subsection{Model Predictive Control (MPC)}

MPC techniques use optimal control framework to solve on-line trajectory generation problems. At each activation, MPC solves an optimal control problem over a sliding time window. In robotics, MPC is usually activated several times per second so algorithms are the same but their applications differ from the off-line trajectories generation. At each activation, a solution is found using only one iteration of the NLP solver. Therefore, the solution used will not be an optimal solution but, by using the previous solution to build the initial guess for the next step, MPC will at the same time improve the solution and adapt it to the new state of the system. The faster the algorithm is, the less the previous solution is out-dated so the more the algorithm will be able to improve it instead of just adapt it to the new situation. Thus, fast single shooting methods like DDP are often used \cite{GaKo2015}. However, DDP is not able to handle constraints, so constraints like obstacle avoidance are introduced in the problem using the cost function and therefore, we cannot be sure that the system will respect those constraints. With the algorithms used in this paper, the linearized constraints are checked at each step so in the case of convex obstacle constraints, we are sure that solutions given by the MPC will always be collision-free (at least at each shooting node).

\section{SYSTEM DYNAMICS} \label{sectionSystem}

In order to exhibit the qualities of direct optimal control to easily adapt to various dynamic models, we have performed a benchmark with three different quadrotor-based robots whose models are given in the three next subsections.

\subsection{Quadrotor}

The quadrotor is modeled as a rigid body of mass $m_{q}=0.9[kg]$ evolving in the 3 dimensional space where effects linked to fluid dynamic are all neglected. Position and orientation of the quadrotor with respect to the inertial frame are respectively noted $\bm{x}_{q} \in \mathbb{R}^{3}$ and $\Theta_{q} \in SO(3)$ (where $\Theta_{q}$ can indistinctly be represented by Euler angles or quaternions) and its rotation speed in its local frame is represented by $\Omega_{q} \in so(3)=\mathbb{R}^{3}$.
Rotor dynamics is neglected so each propeller $i$ produces a thrust $f_{i}=C_{f}V_{i}^{2}$ and a torque around its main axis $\tau_{zi}=(-1)^{i+1}C_{m}V_{i}^{2}$ where $V_{i} \in[V_{min},V_{max}]$ is the motor velocity \cite{model}. Neglecting the rotor dynamics, the control is directly the rotor acceleration. The state vector is $\bm{x} = [\bm{x}_{q},\Theta_{q},\dot{\bm{x}}_{q},\Omega_{q},V_{1},...,V_{4}]^{T}$ and the system control inputs are $u_{i} = \dot{V}_{i}\in[\dot{V}_{min},\dot{V}_{max}]$, $i \in \{1,...,4\}$.

\begin{equation}
m_{q}\ddot{x} = 
\begin{bmatrix}
0 \\
0 \\
-m_{q}g
\end{bmatrix}
+ R_{\Theta_{q}} 
\begin{bmatrix}
0\\
0\\
\sum_{i}C_{f}V_{i}^{2}
\end{bmatrix}
\end{equation}
\begin{equation}
I \Omega_{q} + \Omega_{q} \times (I \Omega_{q}) = 
\begin{bmatrix}
dC_{f}(V_{1}^{2}-V_{3}^{2}) \\
dC_{f}(V_{2}^{2}-V_{4}^{2}) \\
C_{m}(V_{1}^{2}-V_{2}^{2}+V_{3}^{2}-V_{4}^{2})
\end{bmatrix}
\end{equation}
where $d$ is the distance between a rotor and the center of mass of the quadrotor and I is the quadrotor inertia matrix and $R_{\Omega_{q}}$ is the rotation matrix between the world frame an the quadrotor frame.

\subsection{Quadrotor with Pendulum}

The second model corresponds to the same quadrotor, carrying a load attached by a rigid linkage.

\subsubsection{With a fixed load}

We assume that the quadrotor is attached to a load of mass $m_{p}$ with a weightless rigid bar of length $L=4[m]$. The rigid bar and the quadrotor are linked with a spherical joint (3 degrees of freedom). 

To calculate the quadrotor position and the pendulum orientation, the system is modeled as a weightless solid bar with point masses at each tip. Thus, all forces are applied on the axis of the bar and the inertia matrix of the respects $J_{pxx}=J_{pyy}$. Moreover we assume that at the beginning of the trajectory the bar is not rotating around its main axis so Coriolis effects vanish and the dynamic equation can be simplified to
\begin{equation*}
\begin{aligned}
J_{p}\dot{\Omega}_{p} = \sum_{i}T_{i}  && \text{and} &&  \Omega_{pz} = 0
\end{aligned}
\end{equation*}

Thus, the system state vector is $\bm{x} = [\bm{x}_{q}, \Theta_{q}, \dot{\bm{x}}_{g}, \Omega_{q},\theta_{p}, \psi_{p}, \omega_{px}, \omega_{py}, V_{1},..., V_{4}]^{T}$ where $\theta_{p}$ and $\psi_{p}$ are receptively the roll and pitch angles between the world frame and the pendulum frame, $\omega_{px}$ and $\omega_{py}$ the rotation speeds of the pendulum in its local frame and $\dot{\bm{x}}_{g}$ the velocity of the center of mass of the whole system. 

\subsubsection{Grasping and releasing a load}

The mass at the end of the pendulum changes when grasping and releasing. We model the load transfer by a smooth variation of $m_{p}$, to keep the smoothness of the dynamic formulation (discontinuous transfer would have been possible too, but we believe that the smooth transfer more adequately models the linkage). We assume that the velocity of the end-effector with respect to the inertial frame is low at grasping and releasing time so the part of dynamic equation that corresponds to the variation of the mass can be neglected (this assumption is typically correct at the convergence of the direct optimal control solver). 
\begin{equation*}
\dt{p} = m_{q}\ddot{x}_{q} + m_{p}\ddot{x}_{p} + \underbrace{\dot{m}_{p}\dot{x}_{p}}_{\approx 0}  
\end{equation*}
\begin{equation*}
\dt{L_{G}} = GQ \times m_{q}\ddot{x}_{q} + GP \times m_{p}\ddot{x}_{p} + \underbrace{GP \times \dot{m}_{p}\dot{x}_{p}}_{\approx 0}
\end{equation*}
However, when mass of the load changes, the center of mass moves along the pendulum so, at each step, dynamic need to be integrated according to the position/velocity of a new point.
\begin{equation*}
||G_{2}G_{1}|| =  \frac{L m_{q} \dot{m}_{p}}{(m_{q}+m_{p})^{2}+\dot{m}_{p}(m_{p}+m_{q})}dt
\end{equation*}
Where $G_{1}$ is the position of the center of mass at time $t$ and $G_{2}$ is its postion at time $t+dt$. Thus, velocity of the center of mass is computed using
\begin{equation*}
\frac{d\dot{x}_{G}}{dt} = \frac{G_{2}G_{1}}{dt} \times \Omega_{p} + \sum Forces/(m_{p}+m_{q})
\end{equation*}

\begin{figure}%
    \centering
    \subfloat[system quadrotor and pendulum]{{\includegraphics[width=4cm]{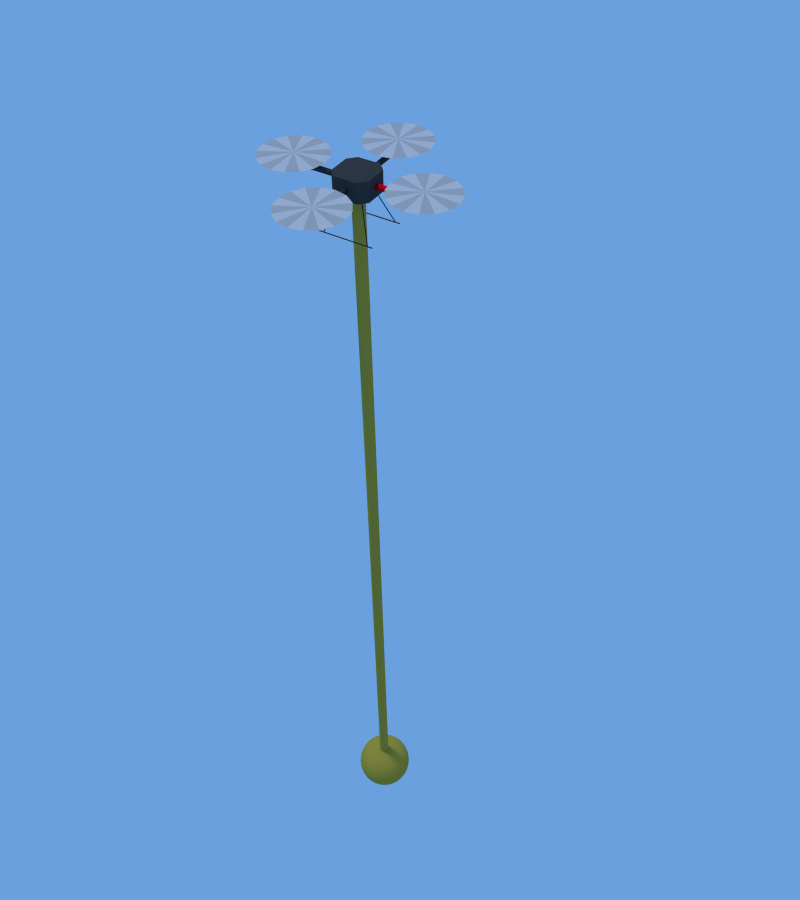} }}%
    \quad
    \subfloat[system quadrotor and arm]{{\includegraphics[width=4cm]{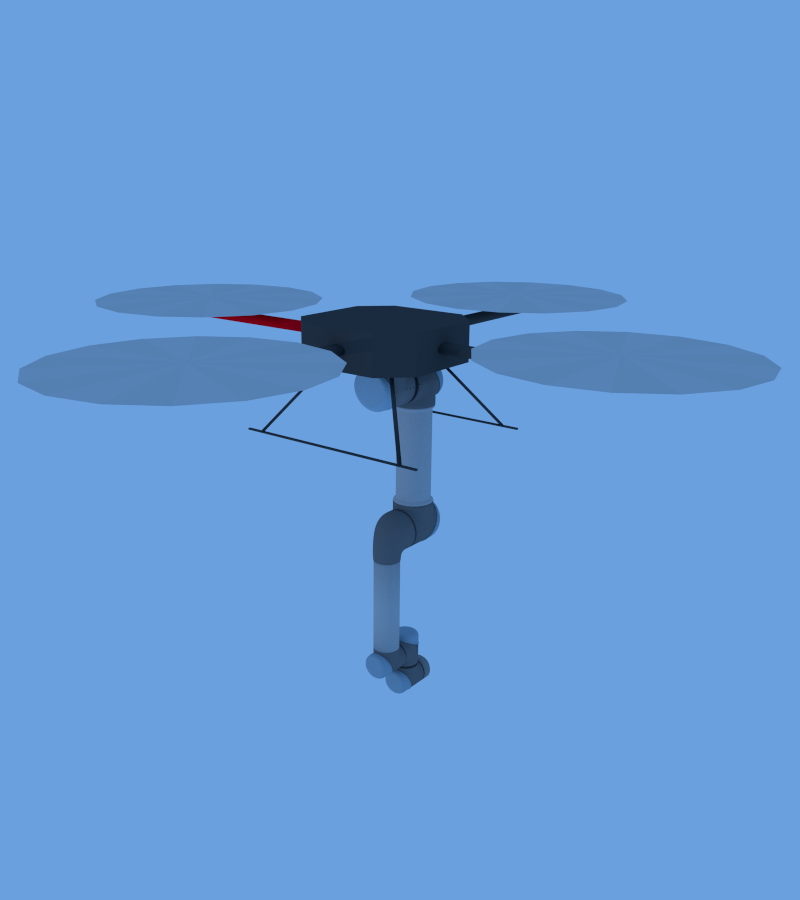} }}%
    \caption{Models of the systems}%
    \label{image:systems}%
\end{figure}

\subsection{Aerial Manipulator}

In that section, we consider that a robotic arm is placed under a quadrotor. The model of the robot arm used for those tests is the \emph{Universal Robot UR5}. For simplicity, the model of the arm is kept as it is and the model of the quadrotor is adapted to correspond about to one that could lift this kind of load (cf. Appendix). Moreover, the quadrotor model is simplified by directly using motor thrusts as controls ($f_{i} \in [f_{min},f_{max}]$, $\tau_{zi} = (-1)^{i+1} C_{m} f_{i}/C_{f}$). The \emph{UR5} has 6 degrees of freedom, its state is the joints positions and velocities while the joint torques are chosen as controls. Therefore, the overall system composed of quadrotor and arm has 24 states and 10 controls. 

While the dynamics of the two previous systems was written algebraically, we rather used for the aerial manipulator an algorithmic form, which is more condensed to write, more versatile and better corresponding to a realistic use. Dynamic of the whole system is calculated using \emph{Recursive Newton-Euler Algorithm} (RNEA) and \emph{Composite Rigid Body Algorithm} (CRBA) \cite{Featherstone}. To speed up calculation, \emph{Pinnocchio} \cite{pinocchio}, a fast implementation of those algorithms is used. From those algorithms, we get
\begin{equation*}
M(q)\ddot{q} + b(\dot{q},q) + g(q) = \tau 
\end{equation*}
where $M(q)$ is the mass matrix (computed by CRBA), $b(\dot{q},q) + g(q)$ are the generalized bias and gravitational forces (computed by RNEA) and $\tau$ is the vector of generalized forces $\tau = [f_{base},\tau_{base},\tau_{joints}]^{T}$ where $f_{base} = [0, 0, \sum_{i}f_{i}]^{T}$, $\tau_{base}$ are the torques induced by the propellers and $\tau_{joints}$ corresponds to the vector of torques acting on each joint of the arm.
To add actuator inertia to this model, a term $M_{mot} = J_{mot}*K_{red}^{2}$ (motor/reducer inertia $J_{mot}$, reduction rate of the reducer $K_{red}$) is added to the mass matrix.
\begin{equation*}
M = M_{crba}+ 
\begin{bmatrix}
 0_{6} & 0_{6} \\
 0_{6} & I_{6}
\end{bmatrix}
M_{mot}
\end{equation*} 
Then, the mass matrix $M$ is inverted (using a sparse cheap algorithm \cite{Featherstone}) to get the direct dynamic needed by the optimal control.
Moreover, constraints on positions and controls are added to the optimization to respect joint limits and maximum torques of the \emph{UR5}.

\section{IMPLEMENTATION DETAILS} \label{sectionImplementationDetails}

\subsection{Initial Guess}

Contrary to single shooting methods where states are implicit variables so the initial guess needs to be specified in the control space, multiple shooting method is initialized within the control and state space. This allows to use simple, noncontinuous but powerful initial guess like quasi-static or obstacle free trajectories and that's why multiple shooting has been shown to be a very useful tool to connect the planning and control parts \cite{MS-Planning}.   

In the experiments, the initial guesses used were mainly quasi-static trajectories between the start and the goal position
\begin{equation*}
\begin{aligned}
&x_{qj} = x_{start}+\frac{j}{N}(x_{goal}-x_{start}),&& \quad j=0,...,N, \\
&V_{i} = \sqrt{mg/4C_{f}},&&\quad i=1,...,4.	
\end{aligned}
\end{equation*}
And all other variables are set to $0$.


\subsection{Obstacle Avoidance}

Contrary to Dynamic Programming where the obstacle avoidance problem is often treated outside then re-injected in the optimal control problem under the form of way-points, direct methods easily deal with constraints on the state space so the obstacle avoidance problem can easily be inserted in the optimal control problem by adding inequalities on the position of the system. 

To get strictly convex and differentiable constraints, obstacles are embedded inside ellipsoids. This model allows to keep trajectories outside of obstacles even when constraints are linearized but also to guide trajectories around obstacles. So in the optimal control problem, the constraint for one obstacle is modeled as
\begin{equation*}
\begin{aligned}
 (\bm{x}_{q}-\bm{x}_{e})^{T} A (\bm{x}_{q}-\bm{x}_{e}) - 1 \ \geq \ 0
\end{aligned}
\end{equation*}
Where $\bm{x}_{q}$ is the quadrotor position, $\bm{x}_{e}$ is the position of the ellipsoid center and $A \in \mathbb{R}^{3 \times 3}$ is a positive definite matrix. 

For task involving obstacles avoidance, a quasi-static trajectory going through obstacles could be a bad initial guess because trajectory continuity and obstacle avoidance constraints can be incompatible when linearized. 
In that case, constraints are relaxed till the obstacle is thin enough to be between two nodes where constraints are checked. From this trajectory, at each step the algorithm will decrease the relaxation term
$u$ in (\ref{relaxation}) 
i.e the obstacle constraint will grows back. When the constraints have finally converged back, the result is usually a trajectory where the system goes fast enough to get through the obstacle when constraints are not checked. Thus, when obstacles are present, the initial guesses used are obstacle free trajectories or at least trajectories where the obstacle constraints does not need to be relaxed.

\subsection{Experimental Setup}

The multiple shooting algorithm has been implemented by Moritz Diehl et al. in an open source optimal control software, the \emph{ACADO toolkit} \cite{acado}. \emph{ACADO} solves multiple shooting problems thanks to a \emph{Sequential Quadratic Programming} (SQP) algorithm, together with state-of-the-art techniques to condense, relax, integrate and differentiate the problem. This tool has already been shown to be useful to generate complex quadcopter trajectories as throwing and catching an inverted pendulum \cite{C2}. Trajectories presented in the following parts have been generated using this toolkit.

Problems are solved on a Intel® Xeon(R) CPU E3-1240 v3 @ 3.40GHz with Runge-Kutta 45 integrator and Broyden-Fletcher-Goldfarb-Shanno (BFGS) Hessian approximation. Moreover, controls are discretized as a piecewise constant function, constant between each shooting node.

\section{RESULTS}

In this section, we report various examples of the capabilities of the optimal control problem solver to discover and control complex trajectories with the three models presented above. The general idea is to give a complete list of the capabilities of the approach in term of range of exploration, speed of computation and robustness to perturbation. For all the reported examples, no external planning methods was use to discover the trajectory (although the aggressive flip trajectory (\ref{sectionTimeOpt}) was not found from scratch by the solver due to its symmetry).
MPC is demonstrated in the case where we were able to set it up. In general, the computation time is sufficient to enable MPC, however in some cases, we were not able to obtain a reasonable trade-off between computation approximation and control performance (in particular for the quadrotor with pendulum). The complete implementation of MPC in all the demonstrated examples along with its application on real quadrotors is left for future works. 

\subsection{Non Optimal Trajectories}

In the case where the task is simple enough, we can encode it using only constraints. Starting with initial guess which does not respect the system dynamic but respects the initial and final constraints (quasi-static trajectory), the algorithm is able to find a valid trajectory after few iterations (convergence criteria : \emph{KKT conditions} under $10^{-12}$). Tab. \ref{ComputationTimeNOCP} shows the computation times for trajectories of 20 shooting nodes over 8 seconds.
\begin{table}
\begin{center}
\begin{tabular}{|p{1.5cm}|p{4cm}|p{2cm}|} 
\hline
System & Task & Time $[s]$ (SQP iterations)\\
\hline
\multirow{3}{1.5cm}{Quadrotor alone} & Horizontal displacement of $10[m]$ & $0.35$ (4)\\
\cline{2-3}
 & Horizontal displacement of $30[m]$ & $0.46$ (5) \\
\cline{2-3}
 & Vertical displacement of $10[m]$ + obstacles avoidance & $0.51$ (6)\\
\hline
\multirow{3}{1.5cm}{Quadrotor + pendulum} & Pendulum stabilization after a perturbation ($\omega_{px\: initial} = 2[rad/s]$) & $0.78$ (4)\\
\cline{2-3}
 & Horizontal displacement of $10[m]$ with inverted pendulum & $0.76$ (4)\\
\cline{2-3}
 & From pendulum $\psi_{p\: initial} = 0$ to inverted pendulum $\psi_{p\: final} = \pi$& $1.99$ (7)\\
\hline
Quadrotor + arm & Reaching a position with the end-effector (distance $5[m]$)& $6.74$ (6) \\
\hline
\end{tabular}
\end{center}
\caption{Computation times for non optimal trajectories.}	
\label{ComputationTimeNOCP}
\end{table}

\subsection{Aggressive Maneuvers}

\subsubsection{Time optimal trajectories with the quadrotor alone} \label{sectionTimeOpt}
Direct methods allow to add parameters to the optimization. In that case, resolution of the optimization will be over the controls and parameters state space. Those parameters can be used to modify dynamics, cost or constraints. So by adding the final time $T$ as parameter to optimize and setting it as cost, the algorithm will try to find a time optimal trajectory respecting the initial and final constraints.
Even if the final time $T$ changed over the SQP iterations, the number of nodes of the multiple shooting remains constant. Therefore, variation of the final time is taken into account when dynamic is computed i.e by the integrator. 
\begin{figure}
\begin{center}
\includegraphics[width=8cm]{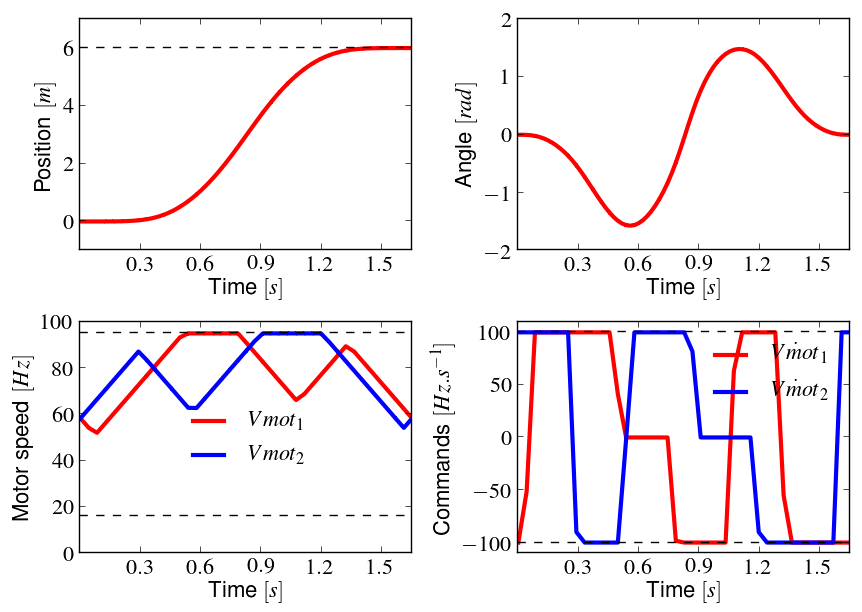} 
\end{center}
\caption{Trajectories for the time optimal problem solved on the quadrotor system.}
\label{TimeOptimal}
\end{figure}
Fig. \ref{TimeOptimal} shows the result obtained for position reaching tasks (6 meters lateral displacement with static terminal constraint) in time optimal with the quadrotor system alone. As we can see, commands calculated by the optimal control are bang null bang as we can expect for a time optimal control. However, the more the algorithm converges to the optimal solution, the more it hits constraints so the longer are the QP resolutions and the smaller are the steps. Therefore, complete convergence is obtained after 57 seconds (373 SQP iterations) although a solution as $T = 1.05 T_{optimal}$ is obtained after only 12.7 seconds (105 SQP iterations). 

When the quadrotor is performing a vertical displacement going up, from a certain distance, the optimal time trajectory is a trajectory where the quadrotor is flipping over at the end of the trajectory and use its propellers to slow down. In \cite{Retournement}, authors showed that an indirect optimal control was able to found those trajectories. Here, the algorithm used is a local algorithm, so it is not able to jump between valid areas of a non-continuous constraints and is only able to find a local optimum. In this case, the end constraint that the system should be in static position is a non-continuous constraint which constrained the final roll and pitch angles to $[\phi_{f},\theta_{f}] = [2k\pi,2l\pi], \quad [k,l]\in\mathbb{N}^{2}$ where $[k,l]$ is actually determined by the initial guess. So, with initial guess used (quasi-static with roll $\phi= 0$ and pitch $\theta=0$), the algorithm is only able to find trajectories where the quadrotor does half a turn in one way then another in the other way instead of a whole turn. Moreover, because of the system symmetry, when the trajectory is strictly vertical, the local optimum is always the trajectory without flip. To find a trajectory with flip, the initial guess needs to be perturb enough to exit the basin of attraction of the solution without flip. The size of the perturbation depend on the length of the trajectory: for a 40 meters high trajectory, setting a lateral displacement of 1 meter in the initial guess is enough to find the flipping trajectory; for less than 20 meters high, a simplified flip needs to be encoded in the initial guess (for instance, for a 20 shooting nodes trajectory, the roll angle is set as $\phi_{i} = \pi$ for $i=\{16, 17, 18\}$ and $\phi_{i}=0$ otherwise).

\subsubsection{Time optimal trajectories with systems quadrotor with pendulum or arm}
Optimal time trajectories for the tasks presented in Tab. \ref{ComputationTimeNOCP} are visible in the attached video.

\subsubsection{Model Predictive Control (MPC)}
To speed up calculation when using the algorithm as MPC, the cost function is simply set as
\begin{equation*}
\begin{aligned}
\int_{0}^{T} &(x_{q}-x_{goal})^{T}C_{1}(x_{q}-x_{goal}) + \Omega_{q}^{T}C_{2}\Omega_{q} \\
\end{aligned}
\end{equation*}
where the term $\Omega_{q}^{T}C_{2}\Omega_{q}$ is used for stabilization of the trajectory and $C_{1}$, $C_{2}$ are weighting matrices. For this test, each 0.2 second a new control is found using the first iteration of a SQP solving a multiple shooting problem of 20 nodes over a 8 seconds sliding time window. The time used to perform one SQP iteration is variable according to the number of QP iterations but here, each SQP iteration took about 0.1 second. This delay is not taken into account in the simulation but on a real application, MPC can use techniques like \emph{delay compensation} \cite{NMPC} to compensate delays between measures of the state and computation of controls.
In the attached video, MPC stability is tested with wind gusts (wind gusts are modeled as a constant piecewise force acting on the quadrotor and no estimation of the perturbation is inserted into the model).

\subsection{Point-to-point Trajectories Through Obstacles}

\subsubsection{Time optimal trajectories through obstacles}
The attached video shows optimal time trajectories with obstacle avoidance for the three systems. 

To simplify constraints with system quadrotor with arm, we suppose that the distance between the obstacle and its constraints is set to be always greater than the length of the arm so the obstacle constraints are only applied on center of mass of the quadrotor.

\subsubsection{MPC}
\begin{figure}
\begin{center}
\includegraphics[width=8cm]{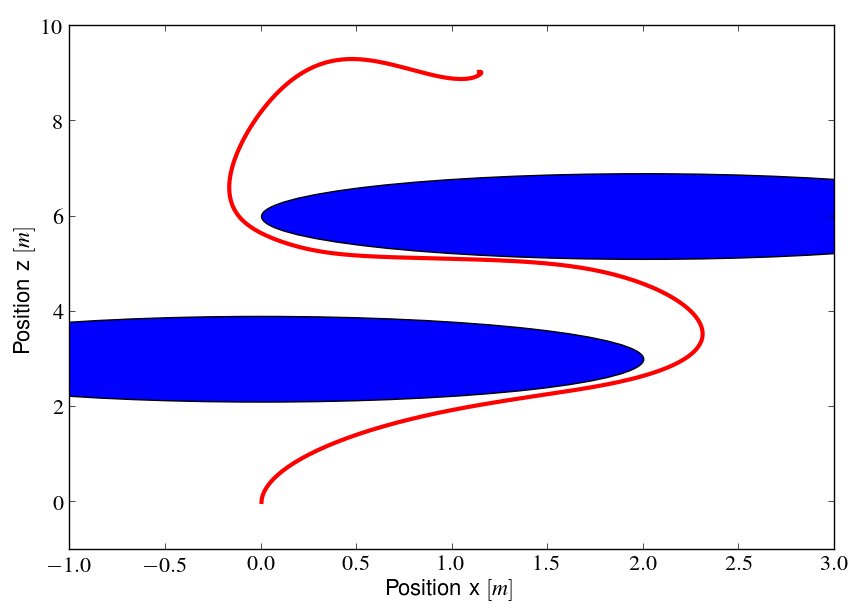} 
\end{center}
\caption{Trajectory of the quadrotor using MPC formulation. The blue ellipses represent the obstacle avoidance constraints.}
\label{MPC1}
\end{figure}

By using simple ellipsoidal constraints for the obstacles, checking those constraints is fast enough to be done online (Fig. \ref{MPC1}).  

\subsubsection{Going through a window}
In this experiment, we ask the quadrotor with pendulum system to go through a window while carrying a load ($m_{p}=0.45$). Since the window height ($2[m]$) is smaller than the pendulum length ($4[m]$), the quadrotor has to find a solution where it swings up the pendulum to get through. The window is supposed infinitely large so only the top and bottom part are modeled as obstacles i.e using ellipsoid constraints. No guess are used (the initial guess is a static trajectory at the initial position) so the algorithm is free to find the best solution according the cost function. The trajectory needs to be able to spread through the window, so no terminal constraints are set on the pendulum state and the terminal constraint that the system should be static is only apply on the quadrotor.

\begin{figure}
\begin{center}
\includegraphics[width=8cm]{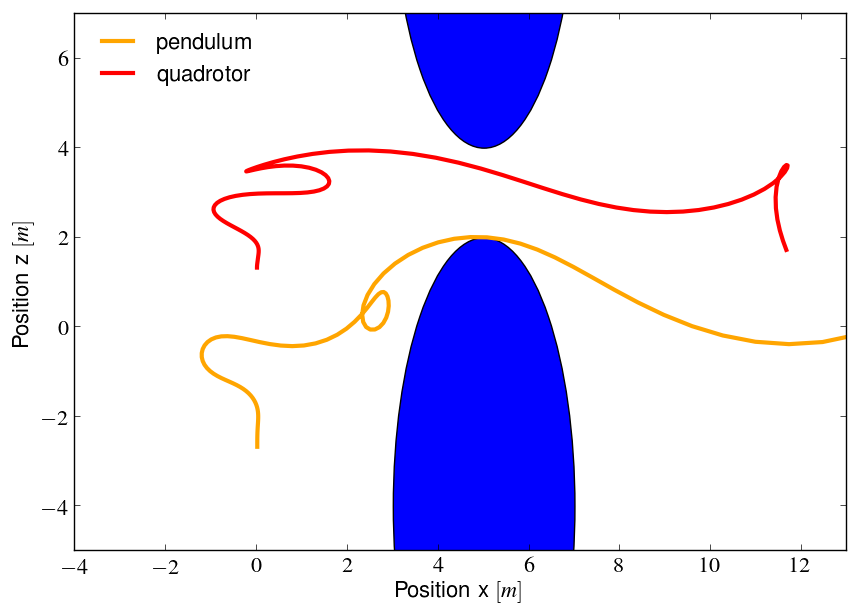} 
\end{center}
\caption{Trajectories of the system quadrotor and pendulum ($m_{q}=0.45[kg]$) while going through a window with terminal cost.}
\label{WindowFrontal}
\end{figure}

In this test, the cost function corresponds to the squared distance between the goal position (which is at the other side of the window) and the quadrotor final position $E(x) = C_{3}||x_{q}(T)-x_{goal}||^{2}$. In this case, the quadrotor starts to move back and forth till the pendulum swings high enough to get through the window (Fig. \ref{WindowFrontal}, computation time : $10.3$ seconds). 

\begin{figure}
\begin{center}
\includegraphics[width=8cm]{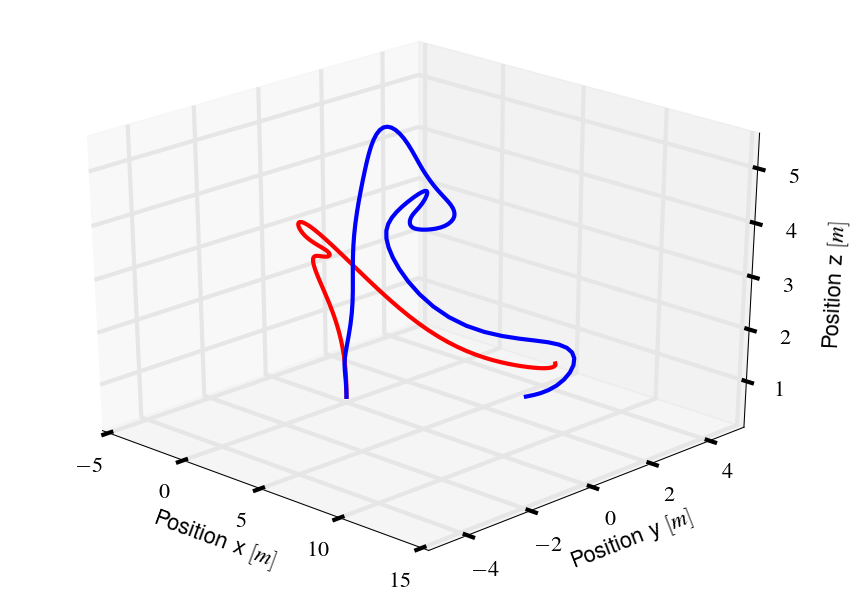} 
\end{center}
\caption{Influence of the cost function on going through a window trajectory. In red, the trajectory using terminal cost function, in blue the one with integral cost function.}
\label{WindowComparaison}
\end{figure}

In the second test, a running cost is added $L(x,u) =  \int_{0}^{T} C_{4}||x_{q}(t)-x_{goal}||^{2} dt$. Here, moving backward at the beginning of the trajectory is costly so instead of moving back and forth, the algorithm is able to find a solution where the quadcopter is moving from left to right to swing the pendulum up. 
(Fig. \ref{WindowComparaison}). 
This kind of behavior wouldn't be possible if we were using waypoints as in \cite{C} because it would restrict the solution to the one encoded by the way-points.

\addtolength{\textheight}{-0cm}   

\subsection{Pick and Place}
In this part, system starts and ends at chosen positions and, on its way, has to pick an object at a certain position then place it to an other one. No obstacles are added to the problem so the object is considered as flying in the air. The task is encoded as follow: the initial and final states are set using initial and final constraints and picking/placing tasks are encoded using the cost function
\begin{align}
L(x,u) = &  \frac{\rho}{\sqrt{2\pi}}e^{\frac{\rho^{2}(t-T_{pick})^{2}}{2}}(\Delta x_{pick}^{T}C_{5}\Delta x_{pick}+\dot{x}_{ee}^{T}C_{6}\dot{x}_{ee}) \label{eq1}\\
+& \frac{\rho}{\sqrt{2\pi}}e^{\frac{\rho^{2}(t-T_{place})^{2}}{2}}(\Delta x_{place}^{T}C_{7}\Delta x_{place}+\dot{x}_{ee}^{T}C_{8}\dot{x}_{ee}) \label{eq2}\\
+& u^{T}C_{5}u \nonumber\\
\Delta x_{pick} &= (x_{ee}-x_{pick}) \nonumber \\
\Delta x_{place} &= (x_{ee}-x_{place}) \nonumber
\end{align}
where $x_{ee}$ is the position of the end-effector, $x_{pick}$ and $x_{place}$ are respectively the positions where the system needs to take and leave the object and $C_{i}$ are weigthing matrices. $\rho$ allows to manage the duration during which the end-effector needs to be at the picking/placing positions and $T_{pick}$, $T_{place}$ are parameters that correspond to the time when object is picked/placed. Terms $\dot{x}_{ee}C_{i}\dot{x}_{ee}$ are used to stabilize the end-effector on the desired position instead of moving around.

\begin{figure}
\begin{center}
\includegraphics[width=8cm]{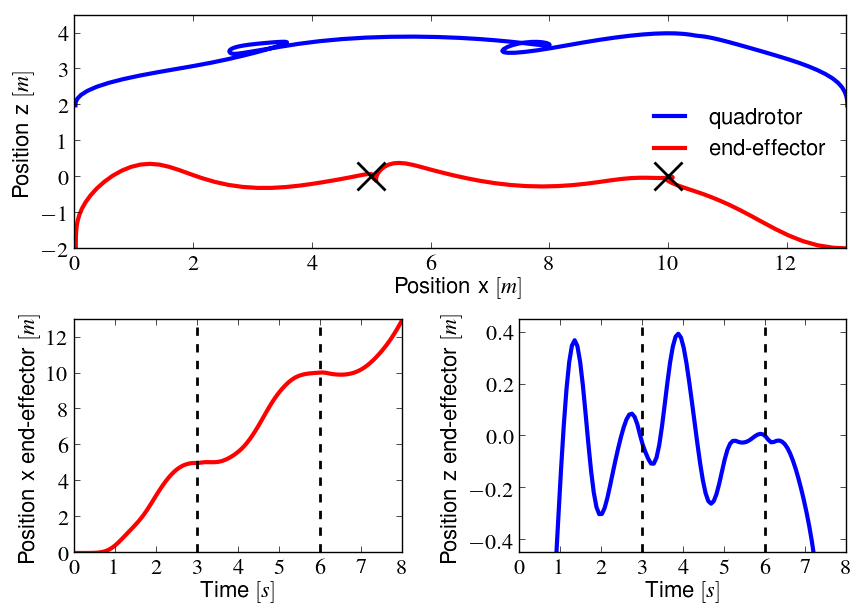} 
\end{center}
\caption{Trajectories of the pendulum for picking and placing task.}
\label{GraphPickPendule}
\end{figure}

For the quadrotor with pendulum, we suppose that the object which needs to be moved has a mass $m_{o}=0.55[kg]$. Therefore, the mass at the bottom of the pendulum will increase from $m_{q}=0.05[kg]$ to $m_{q}=0.6[kg]$ when picking then back to $m_{q}=0.05[kg]$ after releasing the object. So mass is set as a trapezoidal function where the slop lasts one interval of the time grid. Fig. \ref{GraphPickPendule} shows trajectory of the system when picking the object : by swinging the pendulum then drawing an arc of circle around a position, the quadrotor is able to keep the bottom of the pendulum around the same place even if its mass changes. To get a nice behavior like this with a under actuated system, the optimization needs to have enough degrees of freedom so the time grid is set as 60 nodes over a 8 seconds trajectory.
\begin{figure}
\begin{center}
\includegraphics[width=6cm]{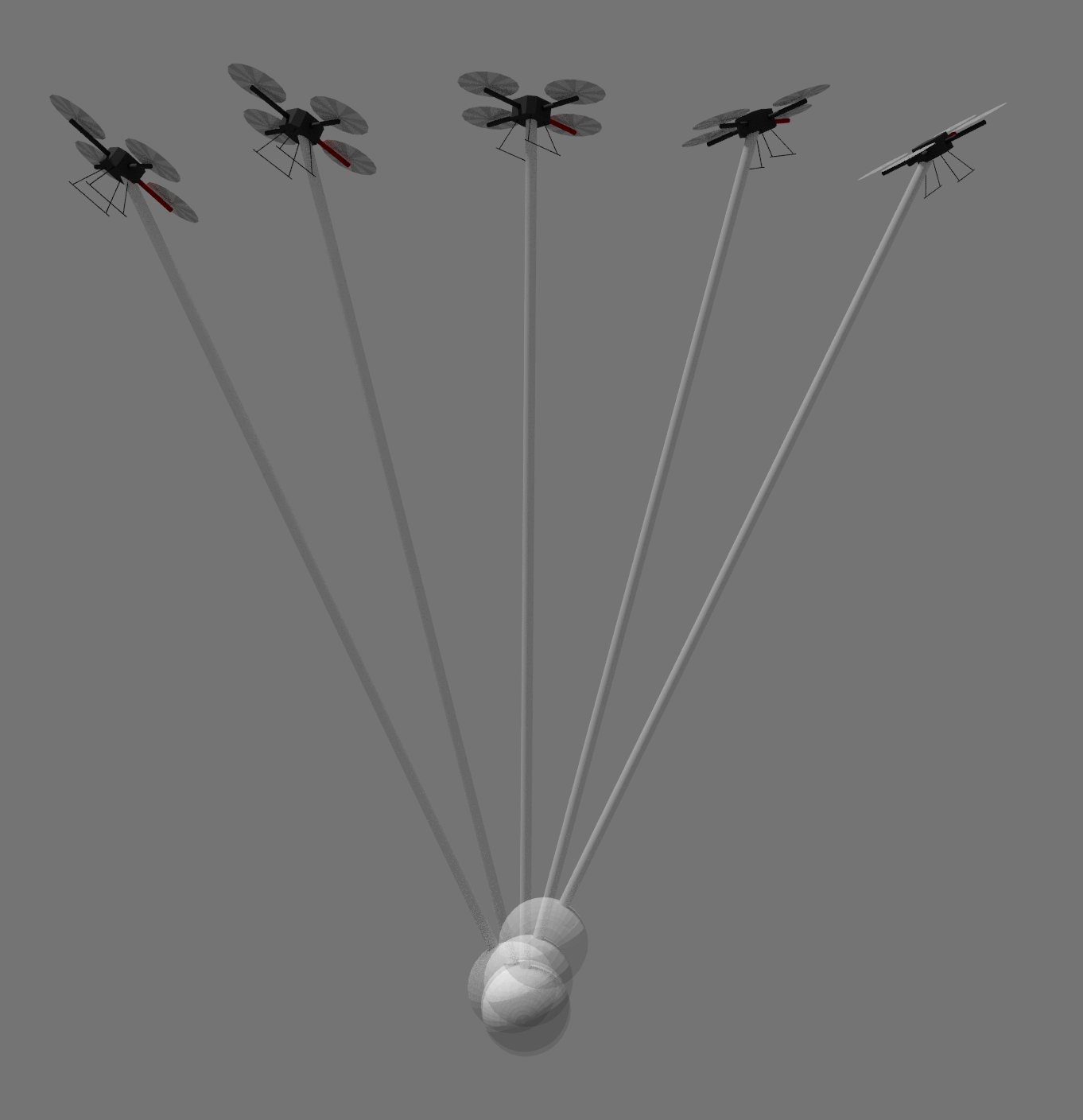} 
\end{center}
\caption{Trajectory of the quadrotor and its pendulum for picking task.}
\label{ImagePickPendule}
\end{figure}

\begin{figure}
\begin{center}
\includegraphics[width=8cm]{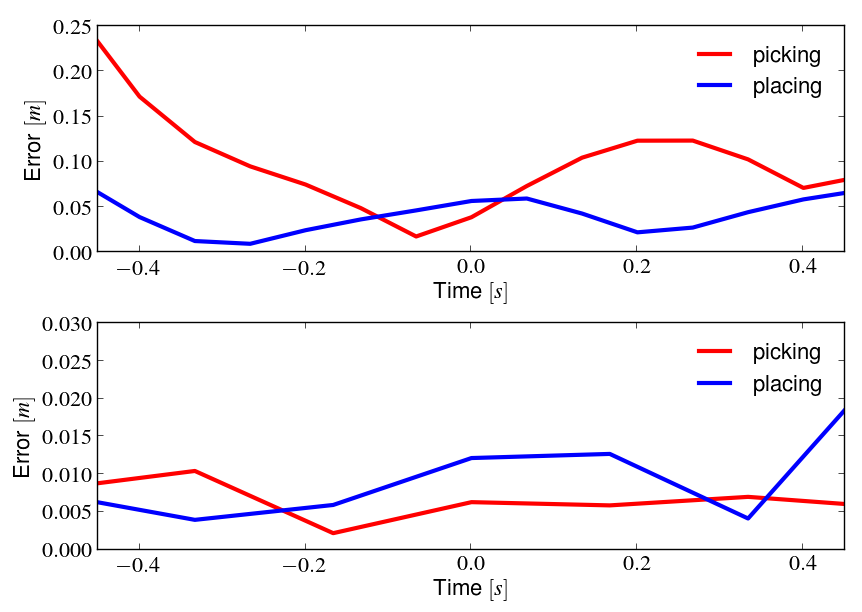} 
\end{center}
\caption{Error between picking/placing position and the end-effector position. At the top, error for system quadrotor and pendulum, at the bottom, error for system quadrotor and arm. Time scale is centered on the picking/placing time.}
\label{ErrorPick}
\end{figure}

\begin{figure}
\begin{center}
\includegraphics[width=8cm]{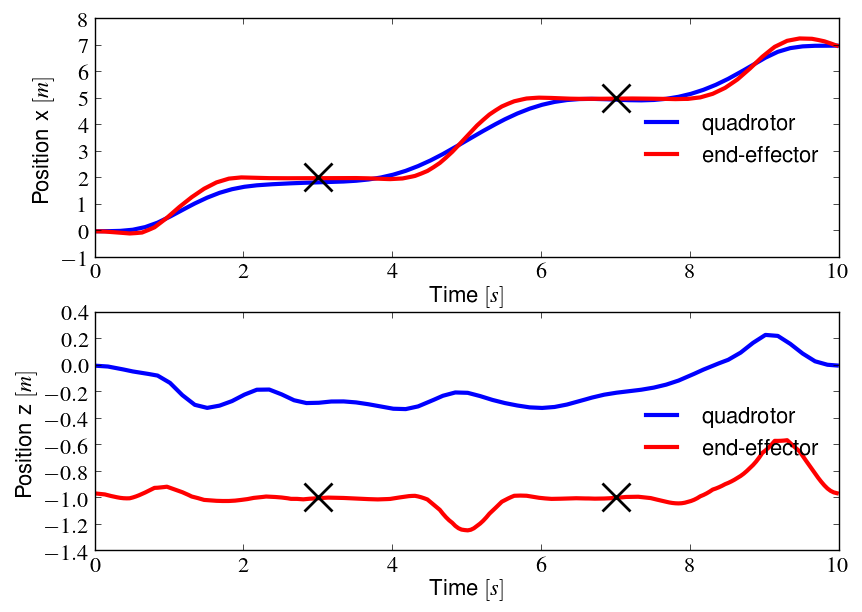} 
\end{center}
\caption{End-effector and quadrotor trajectories for system quadrotor and arm solving the "pick and place" task.}
\label{PickArm}
\end{figure}

For the system quadrotor plus arm, we consider that the object has a weight which is light enough to be neglected. 
Therefore, the system keeps the same dynamic during the whole trajectory and $T_{pick}$ and $T_{place}$ can easily be included in the optimization, so they are set as free parameters with constraints that $0 < T_{pick} < T_{place} < T$ and will be optimized by the algorithm. 
This system has much more actuated degrees of freedom than the last one, thus the algorithm is able to find a end-effector position with a much smaller error (Fig. \ref{ErrorPick}) even with a rougher time grid i.e 20 nodes over a 8 seconds trajectory.
Fig. \ref{PickArm} shows that the algorithm is able to exploit the full dynamic of the system: when performing the picking/placing, the arm is able to compensate for the motion of the quadrotor so it does not need to be in hovering state.
 For those tests, the algorithm does not only solve a task of placing the end-effector at certain positions, at certain times but also tries to keep it as close as possible, as long as possible. Therefore, the shape of the cost function is much more complex and the algorithm needs a lot of iterations to converge a suitable solution (250 SQP iterations so about 7 minutes for both systems).

\subsection{Manipulation Tasks}
In the \emph{Pick and Place} experiment, time varying functions (\ref{eq1}) (\ref{eq2}) are used to specify way-points for the end-effector. For more complex manipulation tasks, the cost function can be used to set a complete reference trajectory for the end-effector. As \emph{Pick and Place}, the optimization needs several minutes to converge to a acceptable solution but allows to exploit the full system dynamic. The attached video shows trajectories where the aerial manipulator is used to drag an object or to turn a crank handle.

\section{CONCLUSIONS}

In this paper, we investigated on how direct methods of numerical optimal control can be exploited to plan and control the motion of complex quadrotor based systems. We empirically demonstrated that this approach is well suited to address several typical problems of UAVs such as generating and achieving aggressive maneuvers, generating oscillatory patterns with a hanging load, dynamically moving through an obstacle field and generating in-contact movements with an aerial manipulator. In particular, we show that multiple-shooting approaches are well suited to discover complex trajectories around obstacles; and we dimensioned the computational load using a prototype implementation (around a few seconds of computations to compute a trajectory from scratch, a tenth of a second to update a trajectory in MPC).
Although the computation cost are very satisfactory, we were not able to set up MPC in a generic manner, apart on the quadrotor alone. This is due to the choice of the cost function, that should regularize the system movements for MPC while we generally setup more aggressive cost for exploration. We will now focus on the construction of cost function more suitable for MPC along with their implementation on physical robots.

\section*{APPENDIX} \label{sectionAppendix}

Quadrotor (values are taken from measures on our own model): $m_{q} = 0.9[kg]$, distance between center of mass and rotor $d = 0.25[m]$, Inertia $J_{q} = diag(0.018, 0.018,0.026)[kg.m^{2}]$, $C_{f} = 6.6 \times 10^{-5}$, $C_{m} = 1 \times 10^{-6}$, $V_{mot_{i}}\in[50,300][rad.s^{-1}]$, $\dot{V}_{mot_{i}}\in[-314,314][rad.s^{-1}]$.
Quadrotor with pendulum : $m_{p} = 0.05[kg]$, $L = 4[m]$.
Aerial manipulator : $m_{q} = 40[kg]$, $J_{q} = diag(10,10,20)[kg.m^{2}]$, distance rotor to center of mass $d=1[m]$, $f_{i}\in[0,200][N]$, \emph{UR5} model taken from its official urdf file, $J_{mot}=5.10^{-6}[kg.m^{2}]$, $K_{red}=250$.  
Window : heigh $h=2[m]$.
Cost functions : $C_{1}=10^{-3}$, $C_{2}=10^{-2}$, $C_{3}=10^{-1}$, $C_{4}=10^{-2}$, $C_{5}=C_{7}=10$, $C_{6}=C_{8}=1$, $\rho=2$


\bibliographystyle{unsrt}
\bibliography{RAL_ICRA_2015.bib}

\end{document}